\documentclass{article} 
\usepackage{iclr2021_conference,times}
\usepackage{amsmath,amsfonts,bm}

\def\eqref#1{equation~\ref{#1}}

\def\1{\bm{1}}

\DeclareMathAlphabet{\mathsfit}{\encodingdefault}{\sfdefault}{m}{sl}
\SetMathAlphabet{\mathsfit}{bold}{\encodingdefault}{\sfdefault}{bx}{n}

\usepackage{hyperref}
\usepackage{url}
\usepackage{graphicx}
\usepackage{xcolor}
\usepackage{soul}

\title{Gaze Perception in Humans and CNN-Based Model}

\author{Nicole X. Han \\
Psychological and Brain Sciences\\
University of California, Santa Barabra\\
\texttt{xhan01@ucsb.edu} \\
\And
William Yang Wang  \\
Computer Science\\
University of California, Santa Barabra\\
\texttt{william@cs.ucsb.edu} \\
\AND
Miguel P. Eckstein  \\
Psychological and Brain Sciences\\
University of California, Santa Barabra\\
\texttt{miguel.eckstein@psych.ucsb.edu} \\
}

\iclrfinalcopy 
\begin{document}
\maketitle

\begin{abstract}
Making accurate inferences about other individuals' locus of attention is essential for human social interactions and will be important for AI to effectively interact with humans. In this study, we compare how a CNN (convolutional neural network) based model of gaze and humans infer the locus of attention in images of real-world scenes with a number of individuals looking at a common location. We show that compared to the model, humans' estimates of the locus of attention are more influenced by the context of the scene, such as the presence of the attended target and the number of individuals in the image.
\end{abstract}

\section{Introduction}

Humans develop as infants the ability to make inferences about other's visual attention  \citep{farroni_eye_2002}. Most studies on social attention present a single face with a gaze cue to the observer \citep{bayliss_orienting_2004,driver_gaze_1999,friesen_eyes_1998,hietanen_social_2002}. The process might seem deceptively simple, but in real-life scenarios, humans need to integrate eye, head orientation, and body posture to make these judgments \citep{azarian_averted_2017,frischen_gaze_2007,hietanen_does_1999}.

Estimating humans' visual focus of attention is also important to allow for more natural human-computer interactions and have applications for early diagnosis of conditions that result in atypical social attention such as autism spectrum disorder \citep{senju_atypical_2009,chawarska_automatic_2003}. There have been numerous developments in computer estimation of gaze or visual focus of attention from images or videos recently \citep{recasens_where_2016,kellnhofer_gaze360_2019,chong_detecting_2020,fischer_rt-gene_2018,zhang_appearance-based_2015}.  

Recent improvements in machine vision showed comparable performance in tasks such as visual recognition as humans. In addition, similar hierarchical processing of visual information between CNNs and human ventral visual cortex in tasks such as object and scene recognition further motivated comparisons between machine and humans \citep{devereux_integrated_2018,cichy_comparison_2016,serre_deep_2019}.
For example, studies have compared humans and deep neural networks across a variety of tasks, including contour detection, visual reasoning with synthetic tasks \citep{firestone_performance_2020}, perception of object silhouettes \citep{pramod_computational_2016}, abstraction and reasoning \citep{chollet_measure_2019}, visual search with real-world scenes \citep{eckstein_humans_2017} or with medical images 
\citep{lago_under-exploration_2021}. However, little is known about how the current gaze estimation models perform compared to humans and their limitations in real-life scenarios.

Here we use a lab-created video dataset with ground-truth about the locus of attention to compare a state-of-the-art convolution neural network-based model \citep{chong_detecting_2020} to human performance. We assess how humans and model errors vary with various image properties to understand the similarities and differences between humans and the model's perception of the locus of attention. Specifically, we ask the following questions: 1. How accurate are humans and models in their estimations? 2. Are human and model errors correlated across images? 3. How does the number of individuals looking at a common spatial location affect humans and the model's estimations? 4. How does the presence of the attended target affect humans and the model? 5. Are humans more consistent in making these estimations?

\section{Methods}
The stimuli used for evaluating human and model performance were image frames extracted from videos that were originally recorded at the University of California, Santa Barbara. Videos included indoor and outdoor scenes such as classrooms, outside campus buildings, dining halls, etc. In each video, multiple students followed an instruction to simultaneously look toward a designated target person. We refer those individuals in the video looking toward the target designated person as gaze-orienting individuals.

\begin{figure*}[!hbt]
    \centering
    \includegraphics[width=\textwidth]{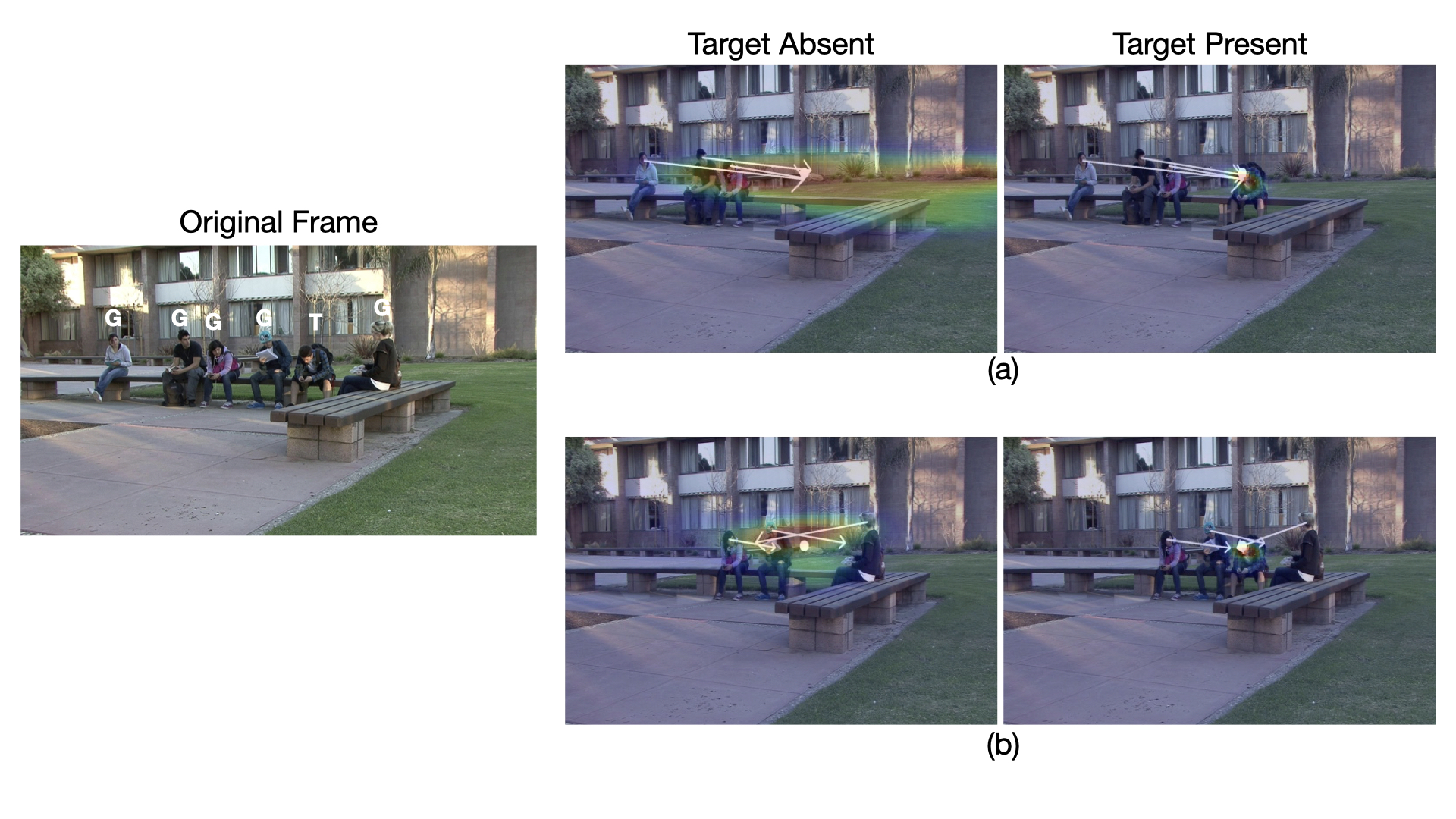}
    \caption{Examples of two versions (a)(b) of stimuli in conditions with designated target present and absent from the same image. "G" indicates the gaze-orienting individuals. "T" indicates the designated target ("attended") individual. The heatmap represents a density map of human estimations of gaze location from the computer mouse selections; arrows represent the model estimation for each gaze-orienting individual, the white dot is the ground-truth location.}
    \label{fig:stimuli}
\end{figure*}

For each video clip, we extracted the image frame in which all the gaze-orienting individuals were looking at the designated target person. We manually segmented each individuals’ body outlines. We randomly selected some gaze-orienting individuals to be deleted during the gaze orienting process. We replaced the RGB values of pixels contained by the body outlines with the RGB values of the immediate background pixels, which can be extracted from other frames in the video. By changing the selected gaze-orienting individuals, we were able to create multiple (2-4) versions of the same image with different gaze-orienting individuals. In addition, we manipulated the number of gaze-orienting people present in the images (one, two, three gaze-orienting individuals). For each image stimuli, we also created a version with a designated target person present in the image (See Figure~\ref{fig:stimuli}).

\subsection{Human behavioral study}

Two hundred participants (at least 18 years old) were recruited with a Human Intelligence Tasks(HIT) posted on Amazon Mechanical Turk. One hundred participants were assigned to the target-present condition, and the other hundred participants were assigned to the target-absent condition. All participants were asked to make an explicit perceptual estimation of the locus of gaze by clicking on the image where they thought all the gaze-orienting individuals were looking at. Sixty images extracted from 60 videos were presented in a random order to each subject (18 images had one gaze-orienting person, 18 images had two gaze-orienting people, 24 images had three gaze-orienting people). Subjects had unlimited viewing time and were able to adjust their selections before moving to the next image. Importantly, each participant only viewed one version of the image from the same video to prevent the memory from interfering with their judgment. 

\subsection{Model Implementation}
We used a state-of-the-art model that was pre-trained on predicting attended targets in videos \citep{chong_detecting_2020}. The model requires the whole image of the scene and a cropped region of one person's head as input. These images are passed to two convolution paths to extract features. For the head region, the model also creates an attention map. After combining scene features, head features, and the attention map, the model uses an Encoder, Conv-LSTM, and Deconv layer to produce a probability map to indicate the predicted attended location in the image. Note that the current version of the model only makes predictions on a single gaze-orienting person at a time.

\section{Results}

\subsection{Data Analysis}
For human estimations of the locus of attention, we used the mean x,y coordinates from all participants' location selections. For the model's estimation, we used the peak location of the probability map for each gaze-orienting individual in the image, and computed their mean location as the model estimation. Note that the current model only takes the original image and just one individual's head region as input rather than multiple head regions at the same time. This is a common limitation in current models and could cause the difference between machines and humans in their ability to integrate context information to perform the task. \citep{chong_detecting_2020}. We also tested other methods to integrate the model estimates across gazing individuals, including the median and the intersection gaze direction vectors pointing to the maximum probabilities in the prediction maps. The mean estimation resulted in higher accuracy.

To calculate human and model estimation error, we calculated the Euclidean pixel distance between the human or model estimation to the designated target person's head location. Section 3.2-3.3 presents data based on target-absent images, while section 3.4 presents data based on both target-absent images and target-present images.

\subsection{Correlation between Humans and Model Perception}

We assessed whether human estimations are more consistent with other humans than when compared to the model. Because most of the variance across images in the attended target location is in the horizontal direction (compared to the vertical direction; the ratio of variance = 18), we calculated the correlation of x-coordinate estimates between human pairs and between human and model pairings. We sampled 10,000 random human pairs and all possible pairings of humans and the model. The mean correlation between humans and the model $(r=0.50)$ was comparable to the mean correlation between pairs of humans $(r=0.55)$ (Appendix Figure~\ref{fig:human_model_error}). 

We also evaluate whether humans tend to be more consistent in their overall estimations across images than the model. With 10,000 bootstrapped samples of human and model errors, we found a significantly smaller variance of human errors ($\sigma^2=1920$) compared to the model ($\sigma^2=6727$), indicating humans have more consistent estimations across different images.

\subsection{Effect of Number of Gaze Orienting Individuals on Estimation Accuracy}
\begin{figure*}[!hbt]
    \centering
    \includegraphics[width=\textwidth]{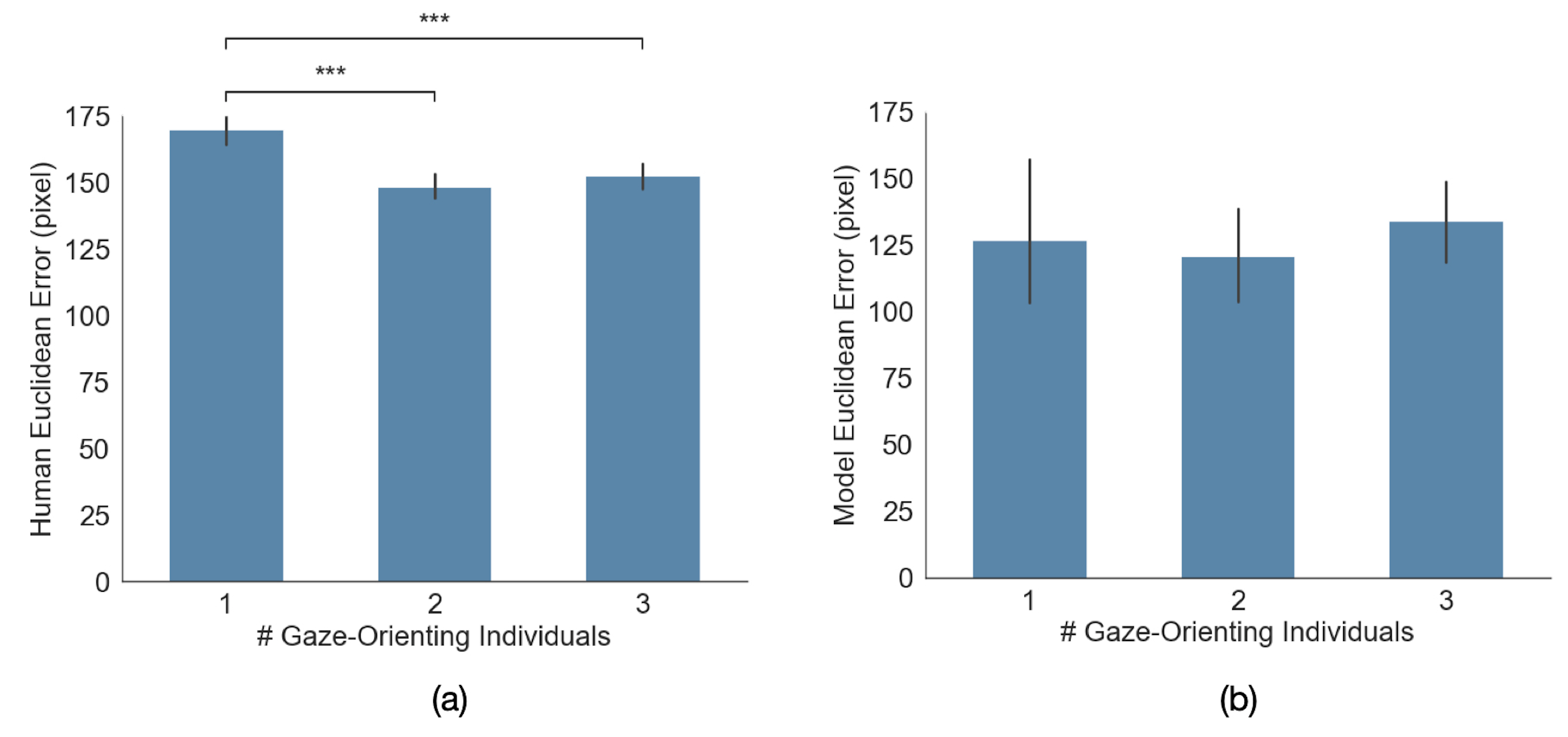}
    \caption{Human and model errors estimating the locus of attention as a function of number of gazing individuals.}
    \label{fig:num_cue}
\end{figure*}

First, we conducted one-way ANOVA to test the effect of the number of looking individuals presented in the image on both humans and the model estimation error. We found a significant main effect of the number looking individuals on human estimation error $F(2,5061)=19.3, p<0.001$, Tukey posthoc test showed a significantly larger error with only one looking individual compared to either images with two or three individuals (both $p<0.001$). In contrast, varying the number of looking individuals in the images did not significantly influence the model's estimation error $F(2,196)=0.5,p=0.58$ (Figure~\ref{fig:num_cue}). A 3 (number of gaze orienting people) $\times$ 2 (human vs. model) two-way ANOVA also confirmed a significant interaction between human and model $F(2,22187)=4.47, p=0.01$. This result indicates that humans are able to integrate multiple gaze cues (individuals) to improve the estimation accuracy, while the model does not benefit from the additional information. 

\subsection{Effect of the Presence of Target}
\begin{figure*}[!hbt]
    \centering
    \includegraphics[width=\textwidth]{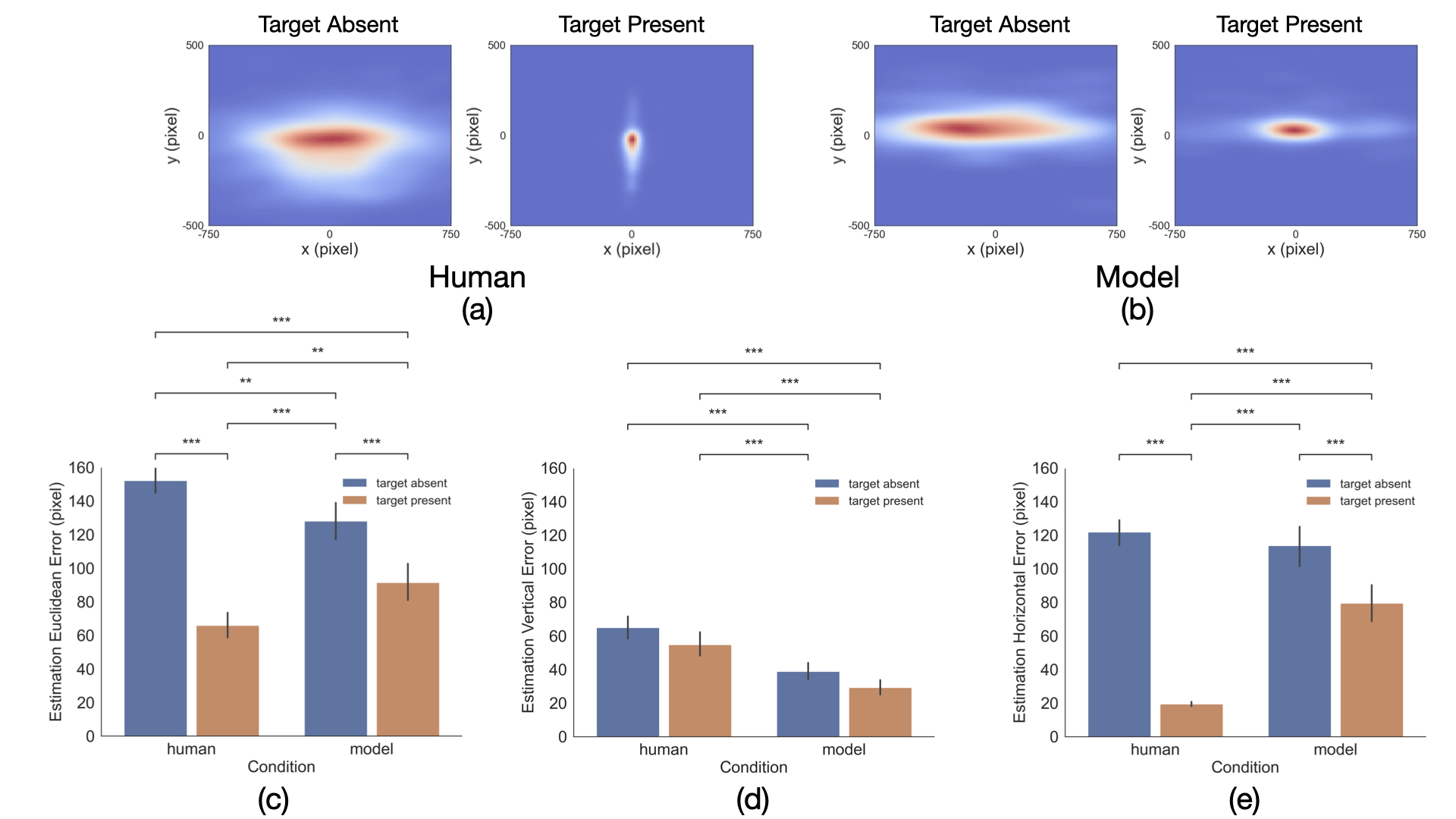}
    \caption{Human and model errors. (a)human and (b) model estimation heatmaps in target-present condition and target-absent conditions, with target location registered at (0,0). (c) overall euclidean estimation error of humans and the model. (d) vertical estimation error. (e) horizontal estimation error.}
    \label{fig:target_presence}
\end{figure*}

We evaluated the humans' and the model's estimation error in instances in which the designated "looked at" target was present vs. absent. Figure~\ref{fig:target_presence}a and 3b show the co-registered locus of attention estimations relative to the location of the target person (origin $(0,0)$), for humans (3a) and model (3b). Two-way ANOVA: estimation source (human vs. model) $\times$ target presence condition (present vs. absent) showed a significant main effect of target presence condition $F(1,690)=134.9,p<0.001$, and a significant interaction between estimation source and target presence condition $F(1,690)=24.2,p<0.001$. A Tukey posthoc test showed that: 1. Both humans and the model had significantly lower errors when the target was present compared to when the target was absent (both $p<0.001$). 2. Humans had significantly higher errors compared to the model when the target was absent, $p<0.01$, but had significantly lower errors when the target was present, $p<0.01$. This indicated that the presence of the attended target in the image facilitates both humans and model to make more accurate judgments about the locus of attention, but with a much larger benefit for humans (Figure~\ref{fig:target_presence}c). 

Results also show a significantly higher amplitude of human estimation errors relative to the model in the vertical direction regardless of the presence of the attended target (Figure~\ref{fig:target_presence}d, all $p<0.001$). For the errors in the horizontal direction, there was no significant difference between humans and the model when the target was absent (Figure~\ref{fig:target_presence}e). However, when the attended target was present, humans errors in the horizontal direction were significantly smaller than the model, $p<0.001$, indicating a strong influence from the presence of the attended target that the model did not show.

\section{Conclusion}
We found a significant correlation between humans and a state-of-the-art CNN-based model's estimates of the locus of attention. Yet, we found some fundamental differences. Humans are more sensitive than the CNN-based model to the scene context. The presence of multiple individuals directing their gaze and the presence of the attended target in the scene reduced the estimation error more greatly for humans. The presence of the attended target person also facilitates humans more than the model in estimation accuracy. Together, the results suggest that humans rely more on multiple sources of scene information to infer social attention than current CNN models. The discrepancy might come from that most of the current models only utilize the whole image and one single head region to estimate gaze. Our findings suggest the necessity of developing models that can integrate more context information and multiple sources of cues in order to achieve better performance in gaze estimation.

\subsubsection*{Acknowledgments}
The research was sponsored by the U.S. Army Research Office and was accomplished under Contract Number W911NF-19-D-0001 for the Institute for Collaborative Biotechnologies. MPE was supported by a Guggenheim Foundation Fellowship.  The views and conclusions contained in this document are those of the authors and should not be interpreted as representing the official policies, either expressed or implied, of the U.S. Government. The U.S. Government is authorized to reproduce and distribute reprints for Government purposes notwithstanding any copyright notation herein.

\bibliography{iclr2021_conference}
\bibliographystyle{iclr2021_conference}

\appendix
\section{Appendix}
\begin{figure*}[!hbt]
    \centering
    \includegraphics[width=\textwidth]{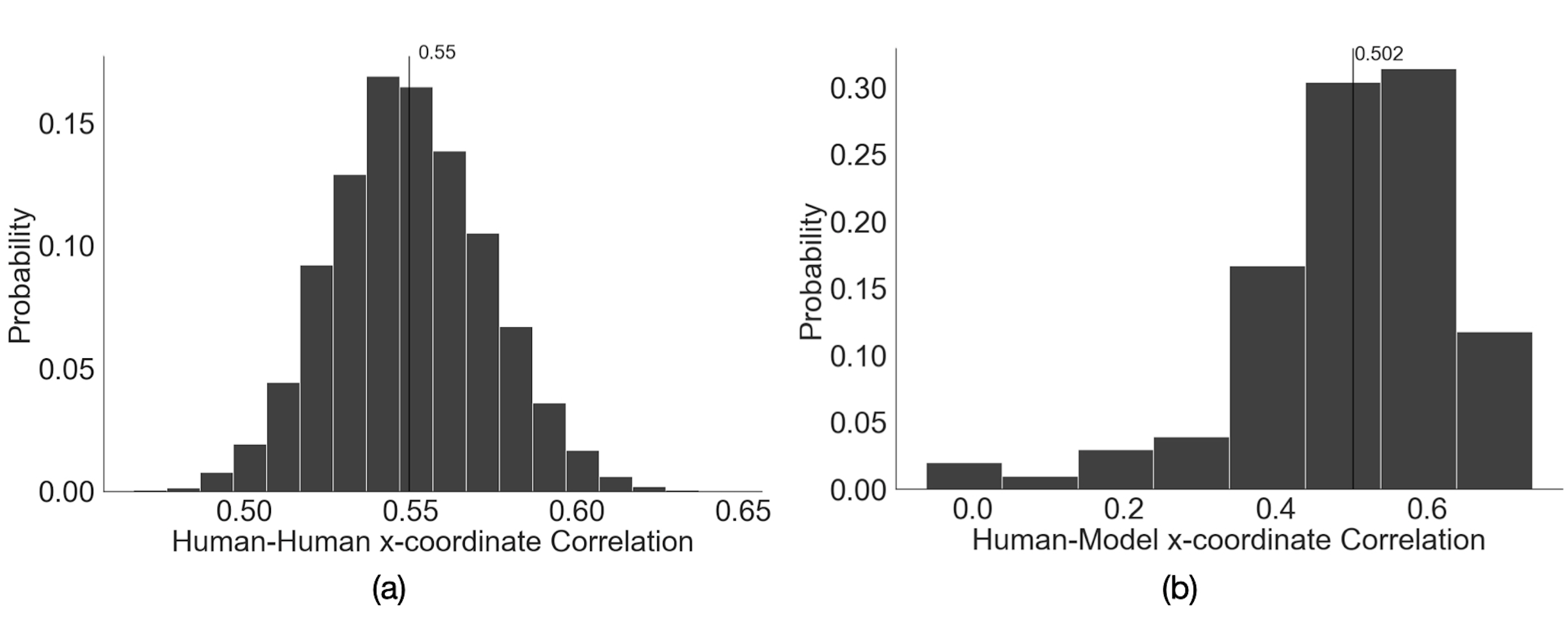}
    \caption{(a) human-human correlation distribution of 10,000 random human pairings. (b) human-model correlation distribution of all pairings of the model and each participant.}
    \label{fig:human_model_error}
\end{figure*}

\end{document}